\documentclass[10pt,twocolumn,letterpaper]{article}

\usepackage{wacv}              

\usepackage{graphicx}
\usepackage{amsmath}
\usepackage{amssymb}
\usepackage{booktabs}

\usepackage{comment}
\usepackage{color}
\usepackage{url}
\usepackage{booktabs, multirow, array, makecell, caption}
\usepackage{verbatimbox}
\usepackage{lipsum}
\usepackage{cuted}
\usepackage{stfloats}
\usepackage{bigstrut} 
\usepackage[table]{xcolor}
\usepackage{rotating}
\usepackage{bm}

\usepackage[accsupp]{axessibility}

\usepackage[pagebackref,breaklinks,colorlinks]{hyperref}

\usepackage[capitalize]{cleveref}
\crefname{section}{Sec.}{Secs.}
\Crefname{section}{Section}{Sections}
\Crefname{table}{Table}{Tables}
\crefname{table}{Tab.}{Tabs.}

\def\wacvPaperID{1400} 
\def\confName{WACV}
\def\confYear{2024}

\begin{document}

\title{Polarimetric PatchMatch Multi-View Stereo}

\author{Jinyu Zhao\\
\and
Jumpei Oishi\\
\and
Yusuke Monno\\
\and
Masatoshi Okutomi\\
Tokyo Institute of Technology\\
{\tt\small \{jzhao, joishi, ymonno\}@ok.sc.e.titech.ac.jp, mxo@ctrl.titech.ac.jp}
}
\maketitle



\begin{abstract}
PatchMatch Multi-View Stereo (PatchMatch MVS) is one of the popular MVS approaches, owing to its balanced accuracy and efficiency. In this paper, we propose Polarimetric PatchMatch multi-view Stereo~(PolarPMS), which is the first method exploiting polarization cues to PatchMatch MVS. The key of PatchMatch MVS is to generate depth and normal hypotheses, which form local 3D planes and slanted stereo matching windows, and efficiently search for the best hypothesis based on the consistency among multi-view images. In addition to standard photometric consistency, our PolarPMS evaluates polarimetric consistency to assess the validness of a depth and normal hypothesis, motivated by the physical property that the polarimetric information is related to the object’s surface normal. Experimental results demonstrate that our PolarPMS can improve the accuracy and the completeness of reconstructed 3D models, especially for texture-less surfaces, compared with state-of-the-art PatchMatch MVS methods.
\end{abstract}

\section{Introduction}


\begin{figure}[t!]
   \centering
   \includegraphics[trim={0cm 0cm 0cm 0cm}, width=1.0\linewidth]{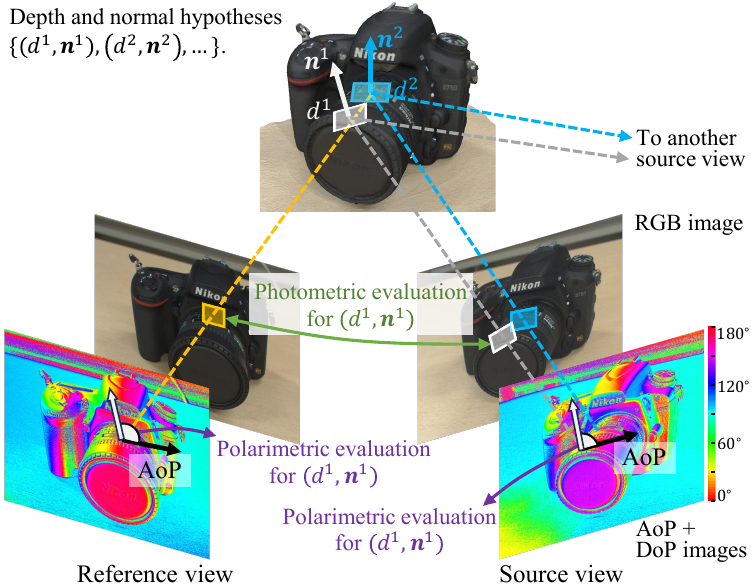}\\
   \caption{For the evaluation of a depth and normal hypothesis, our method evaluates standard photometric consistency and additional polarimetric consistency, which poses a cost to the inconsistency between the observed angle-of-polarization~(AoP) and the azimuth angle of the normal hypothesis.}
   \label{fig:introduction}
\end{figure}

PatchMatch Multi-View Stereo~(PatchMatch MVS)~\cite{bleyer2011patchmatch} is one of the popular MVS approaches
for its good balance in reconstruction accuracy and efficiency~\cite{galliani2015massively,schoenberger2016mvs,wei2014multi,zheng2014patchmatch}. PatchMatch MVS introduces the idea of PatchMatch~\cite{barnes2009patchmatch} to realize an accurate and fast global search for multi-view correspondences. Specifically, PatchMatch MVS methods generally estimate the depth and normal for each pixel of each reference view as follows: (i) Initialize the depth and normal randomly; (ii) Generate depth and normal hypotheses by random generation, perturbation of current estimations, and the propagation considering an adjacent pixel; (iii) Evaluate multi-view consistency to select the best depth and normal hypothesis; (iv) Perform the processes (ii) and (iii) for all the pixels in the scanning order of row-major/column-major and iterate the whole process until sufficient depth and normal results are derived.


In the above processes, a generated depth and normal hypothesis for a pixel of a reference view is used to form a local 3D plane, as shown in the top part of Fig.~\ref{fig:introduction}. This local plane is projected to a source view, which forms a slanted patch for stereo matching. To evaluate the reliability of the depth and normal hypothesis, PatchMatch MVS methods evaluate the photometric consistency (color similarity) between the corresponding patches in the reference and the source views. However, since both the depth and the normal are evaluated based on color textures, it remains challenging to accurately estimate them for texture-less regions.


In this paper, we propose Polarimetric PatchMatch multi-view Stereo (PolarPMS), which is the first method exploiting polarization cues to PatchMatch MVS. Our PolarPMS is motivated by the physical property that the angle-of-polarization (AoP) of reflected light from an object is related to the azimuth angle calculated from the object’s surface normal~\cite{cui2017polarimetric,zhao2022polarimetric}. Based on this, our PolarPMS evaluates polarimetric consistency, which is the consistency between the observed AoP and the azimuth angle from the normal, as shown in the bottom part of Fig.~\ref{fig:introduction}. Compared with existing PatchMatch MVS methods solely based on photometric consistency, our PolarPMS utilizing additional polarimetric consistency mainly has two benefits: (i) Since the AoP is a direct cue to assess the surface normal, the accuracy of the estimated normal map is significantly improved. Furthermore, the improved normal estimation can derive improved depth estimation by effectively incorporating depth-normal consistency evaluation. (ii) Since the AoP is not influenced by the surface texture, the incorporation of polarimetric consistency significantly improves the completeness for texture-less regions. To summarize, the main contributions of this work are as follows:
\begin{itemize}
   \item We propose PolarPMS, which is the first PatchMatch MVS method exploiting polarization information. 
   \vspace{-1mm}
   \item We introduce polarimetric consistency based on AoP, which acts as a direct cue to assess the surface normal regardless of the surface texture, to select the best depth and normal hypothesis in PatchMatch MVS.
   \vspace{-1mm}
   \item We introduce depth-normal consistency to improve the depths by utilizing improved normals with AoP.
   \vspace{-1mm}
   \item We experimentally validate the effectiveness of our PolarPMS, which can derive improved 3D models, especially for texture-less surfaces.
   \vspace{-1mm}
   \item We make the source code publicly available at \url{http://www.ok.sc.e.titech.ac.jp/res/PolarPMS/}.
\end{itemize}

\section{Related Works}
\subsection{PatchMatch Multi-View Stereo}

PatchMatch MVS was originally proposed in~\cite{bleyer2011patchmatch} and its improved versions have been actively proposed~\cite{galliani2015massively,schoenberger2016mvs,wei2014multi,zheng2014patchmatch}.
Inspired by the PatchMatch search algorithm originally designed for image patches~\cite{barnes2009patchmatch}, PatchMatch MVS realizes an accurate and fast global search of the best 3D plane hypothesis that is the most consistent with multi-view input images. However, as common to most other MVS approaches, it remains challenging to derive accurate depth and normal maps for texture-less regions, because PatchMatch MVS methods rely on photometric consistency evaluation to select the best hypothesis.

Some recent PatchMatch MVS methods attempt to address this challenge. TAPA-MVS~\cite{romanoni2019tapa} represents texture-less regions as superpixels and fits a plane for each superpixel.
ACMM~\cite{xu2019multi} applies a multi-scale guided approach, where texture-less regions are considered as better textured when the original images are downsampled.
ACMP~\cite{Xu2020ACMP} and ACMMP~\cite{xu2022multi} combine planar priors to ACMM, which can provide additional geometric constraints to texture-less regions. Even though the plane-based approaches work well for planer scenes, such as buildings and indoor rooms, they may lose surface details if general objects are targeted.  

\subsection{Multi-View Polarimetric Reconstruction}
It is known that the azimuth angle and the zenith angle of the object's surface normal are related to AoP and degree-of-polarization~(DoP) of reflected light from the object~\cite{cui2017polarimetric,zhao2022polarimetric}. Based on these physical properties, many polarimetric 3D reconstruction methods have been proposed in a single-view setting~\cite{smith2018height}, a two-view stereo setting~\cite{atkinson2007shape,fukao2021polarimetric}, or a multi-view setting~\cite{cui2017polarimetric,yang2018polarimetric,zhao2022polarimetric}.



Polarimetric MVS~\cite{cui2017polarimetric} and Polarimetric MVIR~\cite{zhao2022polarimetric} are two closely related methods to our PolarPMS, which adopt multi-view polarization images as the inputs. Polarimetric MVS exploits AoP information to decide the directions to propagate the sparse depths obtained by SfM. While Polarimetric MVS derives dense depth maps even for texture-less regions, the depth propagation is performed in a view-by-view manner without checking multi-view consistency regarding the polarization, leading to limited accuracy. 

Polarimetric MVIR and our PolarPMS utilize a similar observation regarding the relationship between the AoP and the azimuth angle in multi-view consistency. However, they play different roles in the 3D reconstruction steps. PolarPMS is an MVS method that builds a dense point cloud from scratch and uses AoP information to search for the best 3D plane hypothesis in the PatchMatch MVS framework. In contrast, Polarimetric MVIR is a refinement method based on a reasonable initial shape from MVS and uses AoP information for global mesh optimization. Thus, these methods can be applied to each reconstruction step in combination. 



\subsection{Deep-Learning-Based Methods}
Deep-learning-based methods also have been emerging for PatchMatch MVS\cite{wang2021patchmatchnet,lee2021patchmatch,duggal2019deeppruner} and multi-view polarimetric 3D reconstruction~\cite{cao2023multi,tian2023dps}. While these methods demonstrate high potential and performance on their benchmarks, a large amount of training data is still hard to obtain especially for polarization images, for which simulation tools and open datasets are very limited. While a recent method of~\cite{dave2022pandora} based on neural radiance fields does not require the training data, it tends to generate over-smoothed surfaces due to the convolutional nature of the network. Since our PolarPMS builds a 3D model as in a classical MVS approach, it falls into a much different category from those learning-based methods.



\section{Polarimetric PatchMatch Multi-View Stereo}
\label{sec:proposed}

\begin{figure*}[t!]
   \centering
   \includegraphics[trim={0cm 0cm 0cm 0cm}, width=1.0\linewidth]{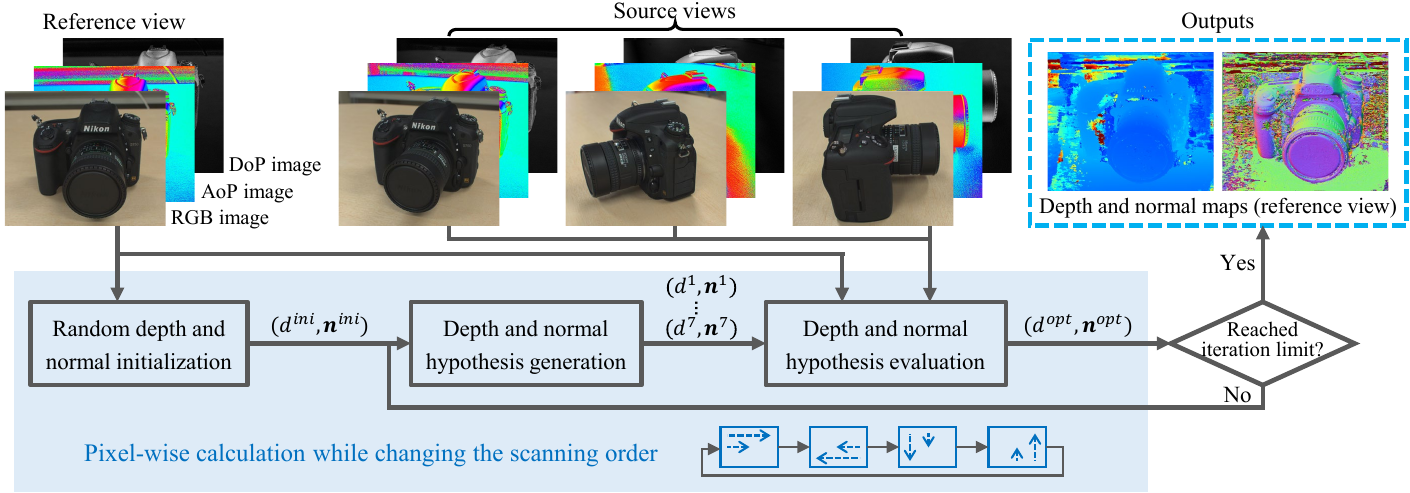}
   \caption{Flow of depth and normal estimation for a reference view in our PolarPMS. According to the PatchMatch algorithm, the depth and normal hypothesis generation and evaluation are performed pixel-wisely in the scanning order of row-major/reverse-row-major/column-major/reverse-column-major at each iteration. We utilize polarimetric information to evaluate the depth and normal hypotheses.}
   \label{fig:flowchart}
\end{figure*}

\subsection{Depth and Normal Estimation Overview}
Our PolarPMS uses multiple polarization images taken from different viewpoints, and camera poses are derived from SfM~\cite{schoenberger2016sfm}.
Figure~\ref{fig:flowchart} shows the flow of depth and normal estimation for a reference view, which is one of the input views. Firstly, the depth and normal for each pixel are randomly initialized. Then, based on the initial/current depth and normal, seven depth and normal hypotheses are generated, as detailed later. Those hypotheses are then evaluated based on the pair-wise consistency between the reference and the source views to select the best hypothesis as the current estimation. This hypothesis generation and selection are performed pixel-wisely in the scanning order of row-major, reverse-row-major, column-major, or reverse-column-major at each iteration until the iteration limit to obtain the final depth and normal estimations. These estimation processes are performed on every input view to obtain the depth and the normal maps of all the views.

\subsection{Depth and Normal Hypothesis Generation}
Given the initial/current depth and normal estimations, seven depth and normal hypotheses are generated in the same way as~\cite{schoenberger2016mvs} as follows.
\begin{equation}
   \label{eq:sampling}
   \begin{aligned}
    \mathcal{A} = \{ &(d_l,\boldsymbol{n}_l),(d^{prop}_{l-1},\boldsymbol{n}_{l-1}),(d^{rnd}_l,\boldsymbol{n}_l),(d_l,\boldsymbol{n}^{rnd}_l),\\
    & (d^{rnd}_l,\boldsymbol{n}^{rnd}_l),(d^{prt}_l,\boldsymbol{n}_{l}),(d_l,\boldsymbol{n}^{prt}_l)
    \},
   \end{aligned}
\end{equation}
where $d_{l}$ and $\boldsymbol{n}_l$ are the current depth and normal of the pixel~$l$, respectively. $d^{prop}_{l-1}$ is the depth obtained by the inter-pixel propagation from the depth and normal of the adjacent pixel $l-1$, which is the depth hypothesis assuming a smooth surface. $d^{rnd}_{l}$ and $\boldsymbol{n}^{rnd}_{l}$ are randomly generated depth and normal hypotheses, and $d^{prt}_{l}$ and $\boldsymbol{n}^{prt}_{l}$ are perturbed hypotheses of the current depth and normal. 

\subsection{Depth and Normal Hypothesis Evaluation}\label{cost}

\subsubsection{Overall Cost Function}
Here, we detail the cost function for depth and normal evaluation. The overall cost function is as follows.
\begin{equation}
\begin{aligned}
    \begin{split}
     (d^{opt}_l\!,\!\boldsymbol{n}^{opt}_l)\!&=\!\arg\min_{d^*_l\!,\!\boldsymbol{n}^*_l}
     \Bigl[
            F_{pho}(d^*_l\!,\!\boldsymbol{n}^*_l)\!+\!\tau_{geo}\!\cdot\! F_{geo}(d^*_l\!,\!\boldsymbol{n}^*_l)\\
            &+\!\tau_{pol}\!\cdot\! F_{pol}(d^*_l\!,\!\boldsymbol{n}^*_l)
            \!+\!\tau_{dep}\!\cdot\! F_{dep}(d^*_l\!,\!\boldsymbol{n}^*_l)
     \Bigr],
     \label{eq:costFunction}
    \end{split}
\end{aligned}
\end{equation}
where $(d^{opt}_{l},\boldsymbol{n}^{opt}_{l})$ is the best hypothesis selected from $\mathcal{A}$ for the pixel $l$ with a minimum cost, and
$(d^*_l,\boldsymbol{n}^*_l)$ represents one of the seven depth and normal hypotheses in $\mathcal{A}$. $F_{pho}$, $F_{geo}$, $F_{pol}$, and $F_{dep}$ evaluate photometric consistency, geometric consistency, polarimetric consistency, and depth-normal consistency, respectively. $\tau_{geo},\tau_{pol}$, and $\tau_{dep}$ are balancing weights. $F_{pho}$ and $F_{geo}$ are based on COLMAP~\cite{schoenberger2016mvs}, and we newly introduce $F_{pol}$ and $F_{dep}$. 
The optimization problem can be solved by Ceres solver~\cite{Agarwal_Ceres_Solver_2022}.

\begin{figure}
\includegraphics[width=1.0 \linewidth]{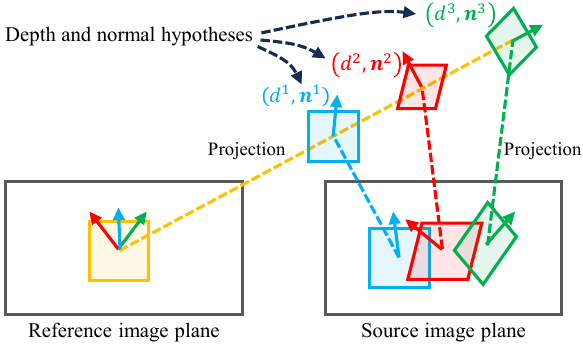}
   \caption{Illustrations of geometric projections used for our cost calculation. }
\label{fig:projection}
\end{figure}

Figure~\ref{fig:projection} illustrates the geometric projections used for cost evaluation. For one depth and normal hypothesis to a considered pixel/patch in the reference image, one local 3D plane hypothesis is generated. This local 3D plane is then projected to the source image plane, by which a slanted patch for photometric consistency evaluation is derived. Similarly, the normal vector is projected to the reference and the source image planes, respectively, which are used to derive the azimuth angle of the normal for polarimetric consistency evaluation. 

\subsubsection{Photometric Consistency}
\label{sec:photoCost}
\begin{figure}
\includegraphics[width=1.0 \linewidth]{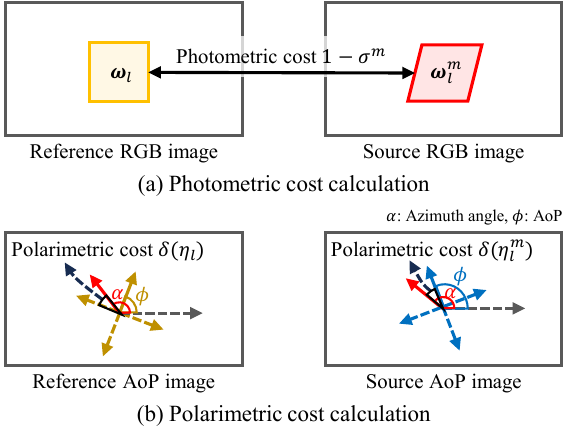}
   \caption{Illustrations of photometric cost calculation and polarimetric cost calculation for hypothesis $(d^2,\boldsymbol{n}^2)$ of Fig.~\ref{fig:projection}.}
\label{fig:cost}
\end{figure}

The photometric consistency between the patch in the reference image and the corresponding patch in the source image (see Fig.~\ref{fig:cost}(a)) is evaluated using standard RGB images as

\begin{equation}
    \begin{aligned}
    F_{pho}(d^*_l,\boldsymbol{n}^*_l)=\frac{1}{|\mathcal{S}|}\sum_{m\in \mathcal{S}}\Bigl[1-\sigma_l^m(d^*_l,\boldsymbol{n}^*_l) \Bigr],
     \label{eq:phoCost}
    \end{aligned}
\end{equation}
where $\mathcal{S}$ is the set of source views, which is adaptively selected for each pixel $l$ of the reference image~\cite{schoenberger2016mvs}, and $m$ is the index of the source view. $\sigma_l^m$ is the color similarity, which is calculated as
\begin{equation}
\begin{aligned}
\begin{split}
\sigma_l^m(d^*_l,\boldsymbol{n}^*_l)=\frac{cov_{\omega}(\boldsymbol{\omega}_l,\boldsymbol{\omega}^m_l)}{\sqrt{{cov_{\omega}(\boldsymbol{\omega}_l,\boldsymbol{\omega}_l)}cov_{\omega}(\boldsymbol{\omega}^m_l,\boldsymbol{\omega}^m_l)}},
 \label{eq:colorSimilarity}
\end{split}
\end{aligned}
\end{equation}
where $cov_\omega$ is the weighted covariance of the patches, $\boldsymbol{\omega}_l$
is the reference patch, and $\boldsymbol{\omega}_l^m$ is the corresponding patch in the $m$-th source image.

This cost function gives a small cost for texture-less regions with similar brightness, regardless of inconsistency in geometry. To avoid such mis-evaluation, COLMAP excludes those regions, where pixel values hardly vary in the reference patch, from the estimation by assigning a maximum cost so as to be removed by later filtering. In contrast, we do not exclude those texture-less regions to improve the completeness by using polarimetric information, as will be explained in Section~\ref{sec:polCost}.

\subsubsection{Geometric Consistency}
The geometric consistency is evaluated as

\begin{equation}
\begin{aligned}
    F_{geo}(d^*_l,\boldsymbol{n}^*_l)=\frac{1}{|\mathcal{S}|}\sum_{m\in \mathcal{S}}\xi^m_l(d^*_l,\boldsymbol{n}^*_l),
 \label{eq:geoCost}
\end{aligned}
\end{equation}
where $\xi$ is robustified geometric cost function:
\begin{equation}
\begin{aligned}
    \xi^m_l(d^*_l,\boldsymbol{n}^*_l)&=1-\sigma^m_l(d^*_l,\boldsymbol{n}^*_l)\\
    &+0.5\min(\psi^m_l(d^*_l,\boldsymbol{n}^*_l),\psi_{max}),
\end{aligned}
\end{equation}

where $\psi^m_l$ represents the distance error between the pixel coordinates of the current pixel and the forward-backward reprojected pixel using the depth value of the reference image (forward direction) and that of the source image (backward direction), and $\psi_{max}$ is the maximum reprojection error, which is set to be three (pixel).

\subsubsection{Polarimetric Consistency}
\label{sec:polCost}

Unpolarized light becomes partially polarized when it is reflected from an object's surface. It is known that the AoP of the reflected light is related to the direction of the surface normal's projection~(i.e., azimuth angle). However, the relationship is not unique and depends on whether specular reflection or diffuse reflection dominates. Generally, there exist $\pi-$ and $\pi/2-$ambiguities in the relationship between the AoP ($\phi$) and the azimuth angle ($\alpha$) without prerequisite knowledge about lighting conditions and surface materials~\cite{zhao2022polarimetric}. This means that four possible azimuth angles ($\phi$, $\phi \pm \pi/2$, $\phi+\pi$) can be inferred from one observed AoP value $\phi$, as illustrated in Fig.~\ref{fig:cost}(b).

Based on the above physical property, we evaluate polarimetric consistency as
\begin{equation}
\begin{aligned}
    F_{pol}(d^*_l\!,\!\boldsymbol{n}^*_l)\!=\!\frac{g(\rho_l)\!\cdot\! \delta(\eta_l(\boldsymbol{n}^*_l)\!)\!+\!\sum\limits_{m\in \mathcal{S}}g(\rho_l^m)\!\cdot\!{\delta(\eta_l^m(d^*_l\!,\!\boldsymbol{n}^*_l)\!)}}{g(\rho_l)\!+\!\sum\limits_{m\in \mathcal{S}}g(\rho_l^m)},
 \label{polCost}
\end{aligned}
\end{equation}
where $\rho_l$ and $\rho_l^m$ represent the DoP values of the reference pixel and the warped pixel in the $m$-th source image, respectively. $g(\rho)$ is the weighting function based on the DoP value, which is detailed later. 

The symbol $\delta$ represents the function of the minimum angle difference $\eta$ between the azimuth angle $\alpha$ of the normal hypothesis $\boldsymbol{n}^*$ and one of the four possible azimuth angles inferred from the observed AoP value $\phi$, as shown in Fig.~\ref{fig:cost}(b). Mathematically, this minimum angle is calculated for the reference pixel $l$ as 
\begin{equation}
\label{eq:nearestAlpha}
\begin{aligned}
\eta_l=\mathop{\min}( & |\alpha_l-\phi_l-2\pi|, |\alpha_l-\phi_l-3\pi/2|,       \\
& |\alpha_l-\phi_l-\pi|, |\alpha_l-\phi_l-\pi/2|, |\alpha_l-\phi_l|,       \\
& |\alpha_l-\phi_l+\pi/2|, |\alpha_l-\phi_l+\pi|).
\end{aligned}
\end{equation}
The minimum angle for the warped pixel in the $m$-th source image $\eta_l^m$ can be calculated in the same way. As the specific function $\delta$, we adopt a concave-shaped function of~\cite{zhao2022polarimetric}, which assigns a higher cost if $\eta$ is larger and vice versa in the range of [0, 1].

As for the weighting, since a high DoP value means high reliability of the polarization information, we assign a larger weight to the view with a larger DoP value as 
\begin{equation}
\begin{aligned}
    g(\rho)=1-[\min(\rho,\rho_0)-\rho_0]^2/\rho_0^2,
 \label{eq:polCost}
\end{aligned}
\end{equation}
which increases when the DoP value $\rho$ becomes larger and reaches the maximum value one when $\rho$ becomes larger than the threshold $\rho_0$. By using this DoP weighting, only reliable AoP values among all the reference and source views are paid attention to assess the polarimetric consistency. Furthermore, if no polarimetric information is available, which means that DoP equals zero for all the views, the polarimetric consistency is neglected, resulting in principally the same result as COLMAP relying on photometric consistency.

\subsubsection{Depth-Normal Consistency}
By introducing the polarimetric consistency, it is expected that the accuracy of the estimated normal is improved. Thus, we aim to affect the improved normal estimations to the depth estimations by enforcing the consistency between the normal and the depth. For this purpose, we introduce a depth-normal consistency evaluation to enforce the consistency between the estimated normal and the normal calculated from the depths in neighboring pixels.
The cost function is calculated as follows.
\begin{equation}
\begin{aligned}
    F_{dep}(d^*_l,\boldsymbol{n}^*_l)=1-(\boldsymbol{n}^*_l)^T\cdot \boldsymbol{n}^{dep}_{l}(d^*_l,d^h_l,d^v_l),
 \label{eq:depthNormal}
\end{aligned}
\end{equation}
where $\boldsymbol{n}^{dep}_{l}$ is the normal of the plane composed of the 3D points associated with the depths of the current pixel $l$~($d^*_l$), its adjucent pixels in the horizontal~($d^h_l$) and the vertical~($d^v_l$) directions.

\subsection{Point Cloud Generation from Multi-Views}
\label{sec:filtering}

After generating the depth and the normal maps of all input views, a point cloud is generated by fusing them. Before the fusion, we apply filtering to remove unreliable estimations. We consider that the estimation is unreliable if a reference pixel/patch $\boldsymbol{x}_l$ does not have both enough polarization cues (DoP value $\rho_l$) and textures (pixel intensity variation in the patch $\lambda_l$).
Thus, we first filter out the pixels that do not satisfy the following conditions.
\begin{equation}
\begin{aligned}
    \mathcal{S}_l=\{\boldsymbol{x}_l|\rho_l \geq \rho_t, \lambda_l \geq \lambda_t\},
\end{aligned}
\end{equation}
where $\rho_t$ is the threshold for the DoP value, and $\lambda_t$ is the variance threshold for the patch, which are set to 0.05 and 1.0 in our paper.
We then follow COLMAP~\cite{schoenberger2016mvs} for the filtering of remaining pixels and the fusion processes, where photometric consistency and geometric consistency are re-checked so that 3D points are generated only from consistent depth and normal estimations in multi-views.

\section{Experimental Results}

\begin{figure}[t!]
   \centering
   \includegraphics[trim={0cm 0cm 0cm 0cm}, width=1.0\linewidth]{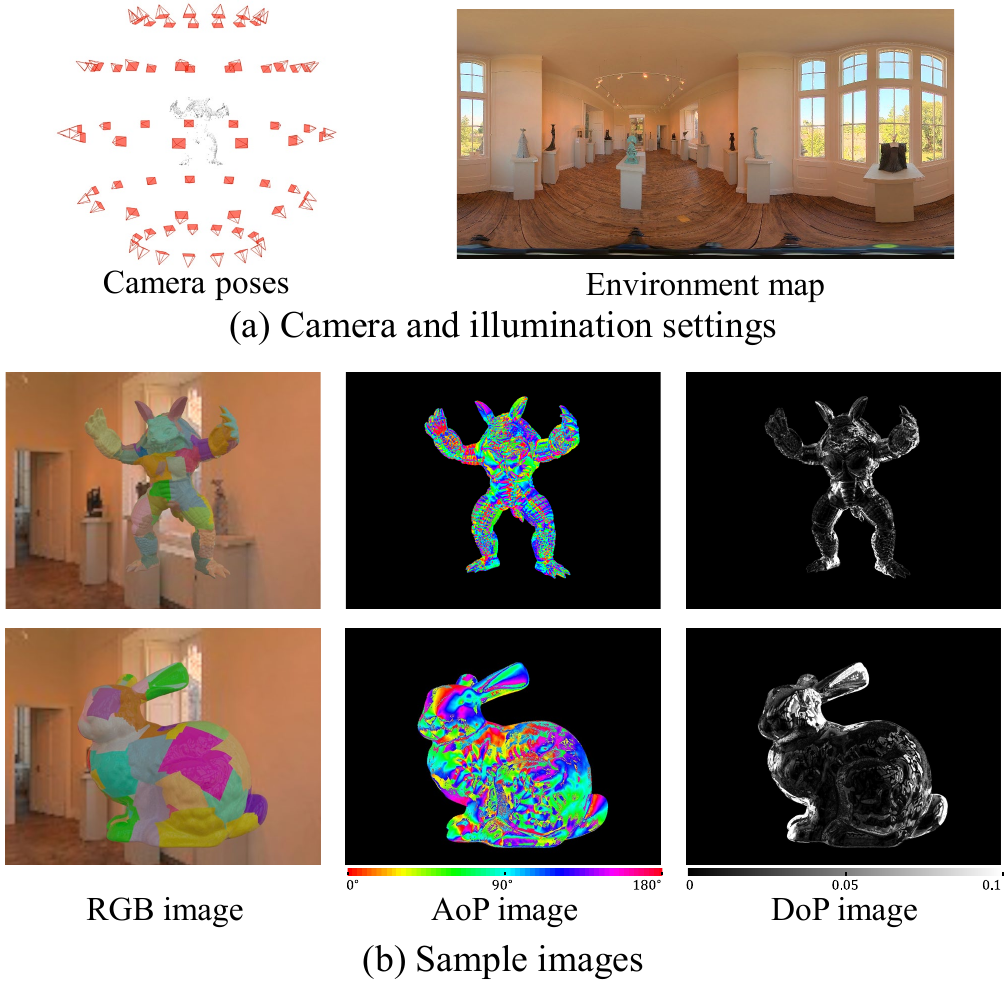}\\
   \caption{Synthetic data generation: (a) Camera and illumination settings; (b) the examples of the synthesized images.}
   \label{fig:setting}
\end{figure}

\subsection{Comparison Using Synthetic Data}
We used three CG models (Armadillo, Bunny, and Dragon) available from Stanford 3D Scanning Repository~\cite{Stanford}. Following~\cite{zhao2022polarimetric}, input RGB, AoP, and DoP images were synthesized using Mitsuba 2 renderer~\cite{nimier2019mitsuba}, which supports a polarimetric BRDF model to simulate realistic polarization images. The synthetic data were generated using a polarized plastic material because it is only the polarized material that can add arbitrary textures. The camera poses and illumination~(environment map) are shown in Fig.~\ref{fig:setting}(a) and the examples of synthesized RGB, AoP, and DoP images are shown in Fig.~\ref{fig:setting}(b).

We empirically set $(\tau_{geo},\tau_{pol},\tau_{dep})$ in Eq.~(\ref{eq:costFunction}) to $(0.4,4.0,0.4)$, and $\rho_0$ in Eq.~(\ref{eq:polCost}) to 0.005,
and compared our PolarPMS with COLMAP~\cite{schoenberger2016mvs}, which is the base of our method, and ACMMP~\cite{xu2022multi}, which is a state-of-the-art PatchMatch MVS method incorporating planer priors.  

\begin{figure}[t]
   \centering
   \includegraphics[trim={0cm 0cm 0cm 0cm}, width=1.0\linewidth]{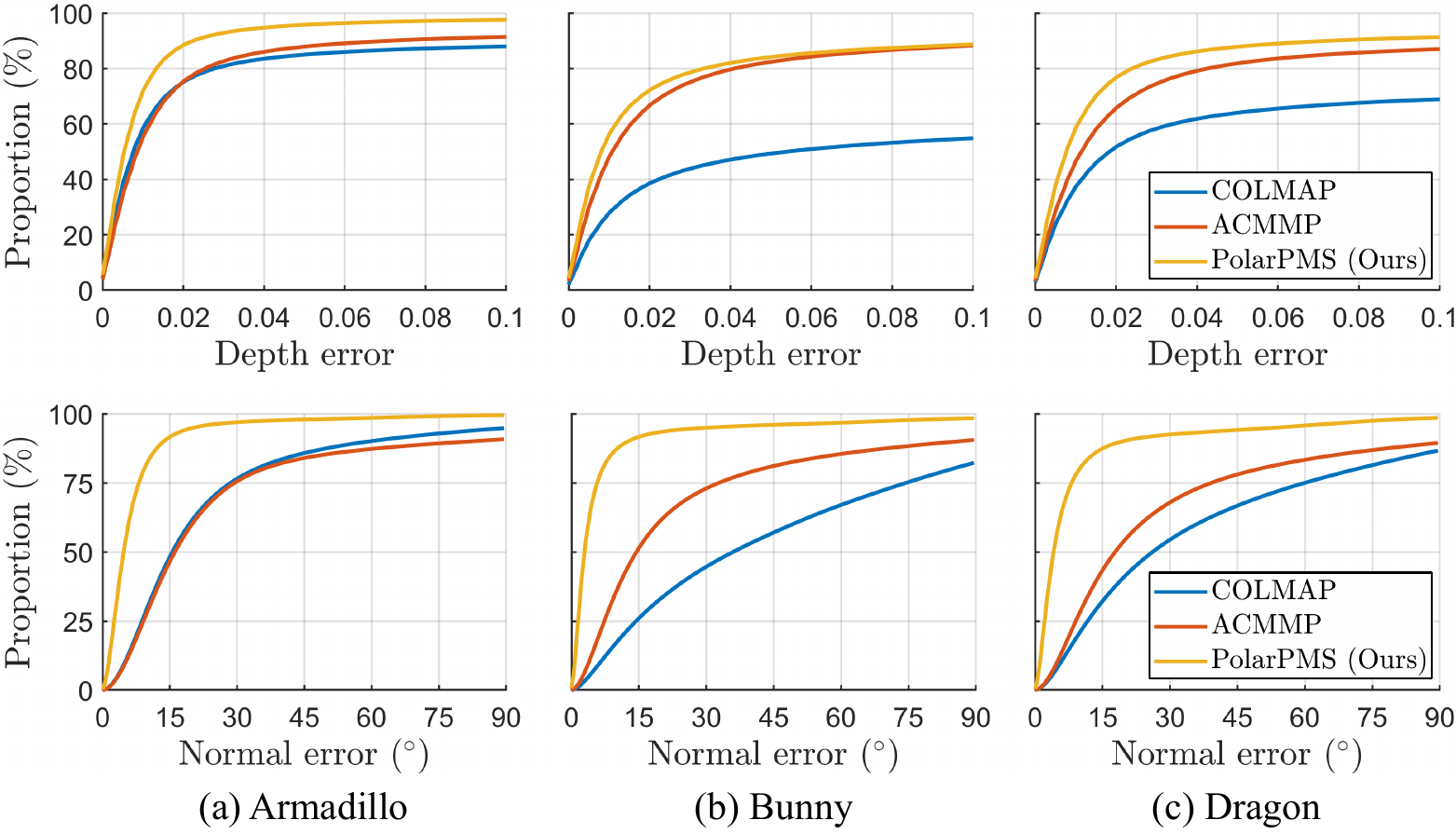}\\
   \vspace{-1mm}
   \caption{Numerical evaluation of depths (top) and normals (bottom) of all pixels. The vertical axis represents the proportion of the pixels whose errors are less than the threshold of the horizontal axis.}
   \label{fig:simulation_plot}
\end{figure}

\begin{figure}[t]
   \centering
   \includegraphics[trim={0cm 0cm 0cm 0cm}, width=1.0\linewidth]{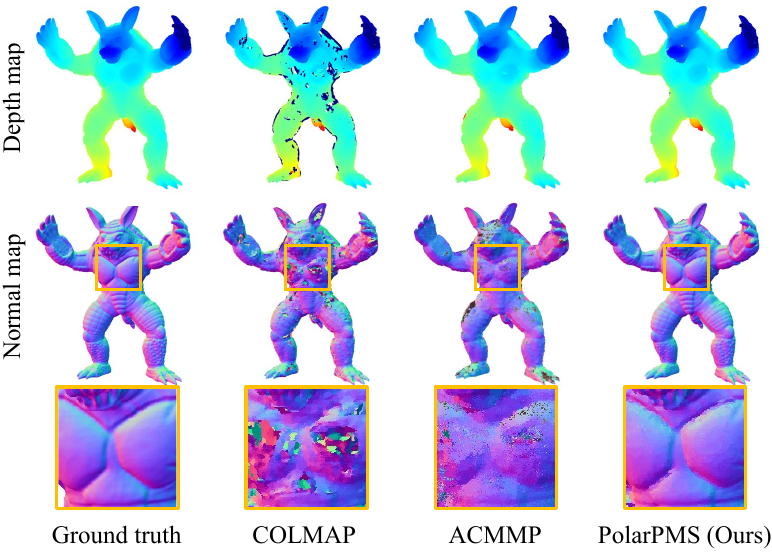}\\
   \caption{Depth and normal map comparison.}
   \label{fig:simulation_map_depth}
\end{figure}

Figure~\ref{fig:simulation_plot} numerically evaluates the depth and normal estimation results of all the pixels of all the views, where $x$-axis indicates the depth or normal error threshold, and $y$-axis represents the proportion of the pixels with a smaller error than the threshold. We can clearly see that our PolarPMS achieves the best accuracy among compared methods with the fastest ascent speed and the highest proportion of the pixels with less error than a certain threshold. Especially, the normal accuracy is significantly improved, owing to our polarimetric consistency evaluation that directly affects the selection of reliable normal hypothesis. This improvement can be seen visually in Fig.~\ref{fig:simulation_map_depth}, where our PolarPMS derives significantly better normal results.

\begin{figure}[t]
   \centering
   \includegraphics[trim={0cm 0cm 0cm 0cm}, width=1.0\linewidth]{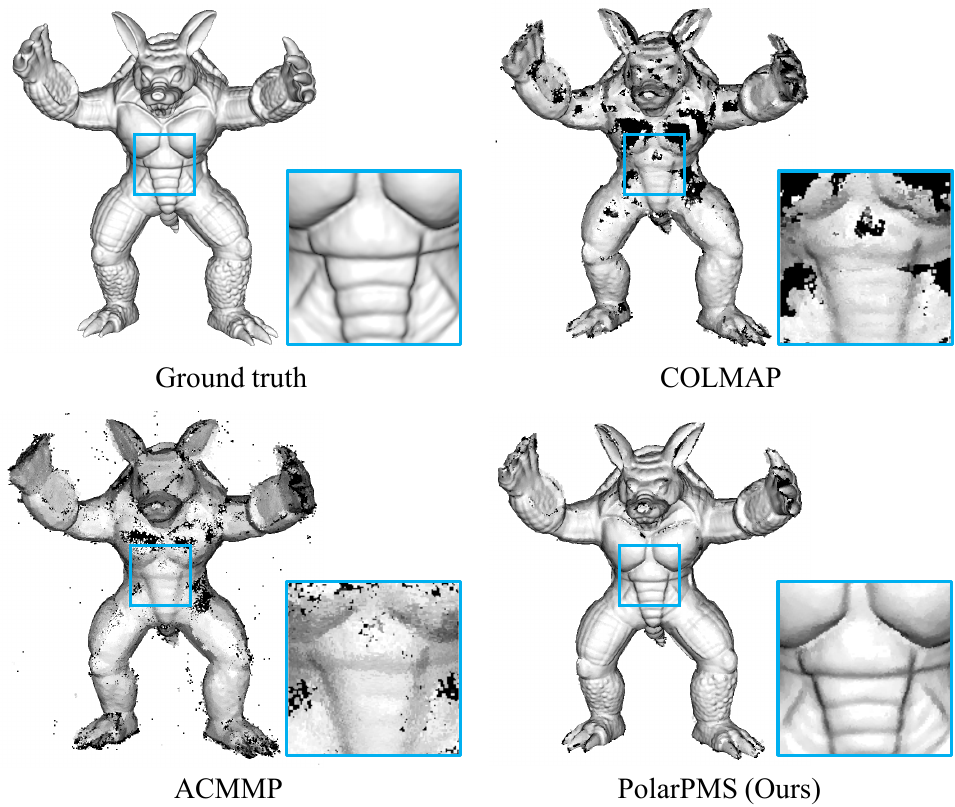}
   \caption{Visual comparison for the Armadillo model.}
   \label{fig:simulation_shape}
\end{figure}

\begin{figure}[t]
   \centering
   \includegraphics[trim={0cm 0cm 0cm 0cm}, width=1.0\linewidth]{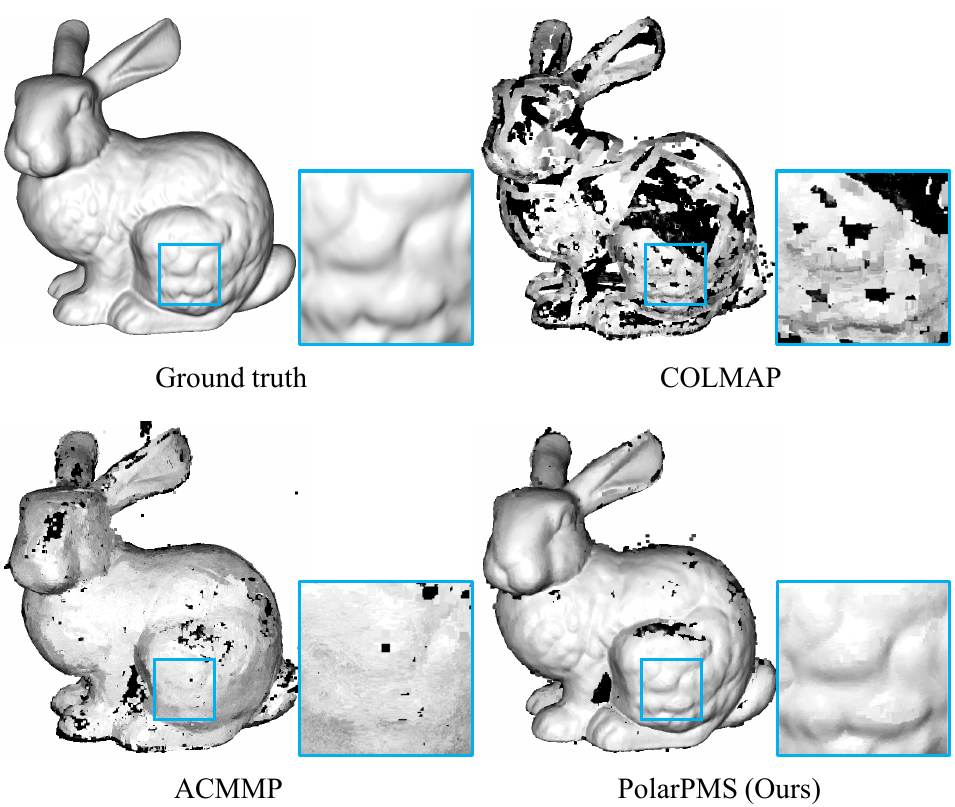}
   \caption{Visual comparison for the Bunny model.}
   \label{fig:simulation_shape_bunny}
\end{figure}

We then compare 3D point cloud results derived from depth and normal maps, using the same fusion parameters for all the compared methods.
The quality of the reconstructed point is evaluated using two commonly-used metrics~\cite{aanaes2016large,ley2016syb3r}: ``Accuracy" which is the distance from each estimated 3D point to its nearest ground-truth 3D point and ``Completeness" which is the distance from each ground-truth 3D point to its nearest estimated 3D point. Table~\ref{table:evaluation} summarizes the average accuracy and completeness for each model. Figures~\ref{fig:simulation_shape} and~\ref{fig:simulation_shape_bunny} show the visual comparisons for the Armadillo model and the Bunny model, respectively. Generally, COLMAP achieves good accuracy, but poor completeness, because COLMAP only focuses on the regions with rich texture and excludes the texture-less regions from the estimation due to the difficulty in photometric consistency evaluation. This is extremely conspicuous for the Bunny model of Fig.~\ref{fig:simulation_shape_bunny}. ACMMP aims to reconstruct texture-less regions and exhibits the effectiveness by better completeness than COLMAP. However, ACMMP tends to lose the surface details because it assumes local planer scenes, which is not always the case, as shown in the enlarged regions of Figs.~\ref{fig:simulation_shape} and~\ref{fig:simulation_shape_bunny}.
In contrast, our PolarPMS achieves good performance in terms of both accuracy and completeness, thanks to our polarimetric consistency evaluation, which is effective regardless of surface textures, to derive better surface normals and depth-normal consistency evaluation to utilize the improved normals to improve depths too. 

\begin{table}[t]
   \centering
   \renewcommand\arraystretch{1.2}
   \setlength\aboverulesep{0pt}\setlength\belowrulesep{0pt}
   \setcellgapes{1.5pt}\makegapedcells
   \caption{Comparisons of the average accuracy (Acc.) and completeness (Comp.) errors}
   \label{table:evaluation}
\scalebox{0.73}{
\begin{tabular}{l|p{6em}|l|l|l}
\toprule
\multicolumn{1}{l}{} & \multicolumn{1}{l|}{} & \multicolumn{1}{p{4.5em}|}{COLMAP} & \multicolumn{1}{p{4.5em}|}{ACMMP} & \multicolumn{1}{p{7em}}{PolarPMS (Ours)} \\
\midrule
\midrule
\multirow{3}[6]{*}{Armadillo} & \multicolumn{1}{l|}{\# of Points} & 694,448  & 1,110,869  & 918,007  \\
\cmidrule{2-5}      & Acc.($\times10^{-2}$) & 0.622  & 0.679  & \textbf{0.522 } \\
\cmidrule{2-5}      & Comp.($\times10^{-2}$) & 1.035  & 0.835  & \textbf{0.568 } \\
\midrule
\multirow{3}[6]{*}{Bunny} & \multicolumn{1}{l|}{\# of Points} & 385,511  & 1,231,652  & 1,192,849  \\
\cmidrule{2-5}      & Acc.($\times10^{-2}$) & \textbf{0.689 } & 0.780  & 0.778  \\
\cmidrule{2-5}      & Comp.($\times10^{-2}$) & 5.373  & 1.483  & \textbf{0.867 } \\
\midrule
\multirow{3}[6]{*}{Dragon} & \multicolumn{1}{l|}{\# of Points} & 538,665  & 1,040,791  & 1,117,226  \\
\cmidrule{2-5}      & Acc.($\times10^{-2}$) & 0.714  & 0.799  & \textbf{0.669 } \\
\cmidrule{2-5}      & Comp.($\times10^{-2}$) & 3.894  & 2.572  & \textbf{1.635 } \\
\midrule
\midrule
\multirow{2}[4]{*}{Average} & Acc.($\times10^{-2}$) & 0.675  & 0.753  & \textbf{0.656 } \\
\cmidrule{2-5}      & Comp.($\times10^{-2}$) & 3.434  & 1.630  & \textbf{1.023 } \\
\bottomrule
\end{tabular}%
}
 \end{table}%

We also performed an ablation study to confirm the effectiveness of our proposed polarimetric and depth-normal consistencies. Table~\ref{table:albation} summarizes the average pixel-wise depth and normal errors. Compared to the first column where neither polarimetric consistency nor depth-normal consistency is introduced~(principally COLMAP), the second column and the third column demonstrate that the accuracy of both depth estimation and normal estimation can be improved by considering polarimetric and depth-normal consistencies, respectively. The fourth column shows that the best results can be derived by introducing polarimetric and depth-normal consistencies simultaneously.

\begin{table}[t]
   \centering
   \renewcommand\arraystretch{1.2}
   \setlength\aboverulesep{0pt}\setlength\belowrulesep{0pt}
   \setcellgapes{1.5pt}\makegapedcells
   \caption{Comparisons of the average depth and normal errors}
   \label{table:albation}
\scalebox{0.83}{
\begin{tabular}{l|l|r|r|r|r}
\toprule
\multicolumn{2}{l|}{Polarimetric consistency} &       & \multicolumn{1}{c|}{\checkmark} &       & \multicolumn{1}{c}{\checkmark} \\
\midrule
\multicolumn{2}{l|}{Depth-normal consistency} &       &       & \multicolumn{1}{c|}{\checkmark} & \multicolumn{1}{c}{\checkmark} \\
\midrule
\midrule
\multirow{2}[4]{*}{Armadillo} & Depth & 0.066  & 0.034  & 0.049  & \textbf{0.026 } \\
\cmidrule{2-6}      & Normal (deg) & 24.429  & 8.897  & 18.344  & \textbf{7.845 } \\
\midrule
\multirow{2}[4]{*}{Bunny} & Depth & 0.338  & 0.155  & 0.346  & \textbf{0.103 } \\
\cmidrule{2-6}      & Normal (deg) & 46.348  & 10.430  & 40.707  & \textbf{8.054 } \\
\midrule
\multirow{2}[4]{*}{Dragon} & Depth & 0.188  & 0.113  & 0.167  & \textbf{0.082 } \\
\cmidrule{2-6}      & Normal (deg) & 37.309  & 11.864  & 29.956  & \textbf{10.263 } \\
\midrule
\midrule
\multirow{2}[4]{*}{Average} & Depth & 0.197  & 0.101  & 0.187  & \textbf{0.070 } \\
\cmidrule{2-6}      & Normal (deg) & 36.029  & 10.397  & 29.669  & \textbf{8.721 } \\
\bottomrule
\end{tabular}%
}
 \end{table}%
 
\subsection{Comparison Using Real Data}
\begin{figure*}[t]
   \centering
   \includegraphics[trim={0cm 0cm 0cm 0cm}, width=1.0\linewidth]{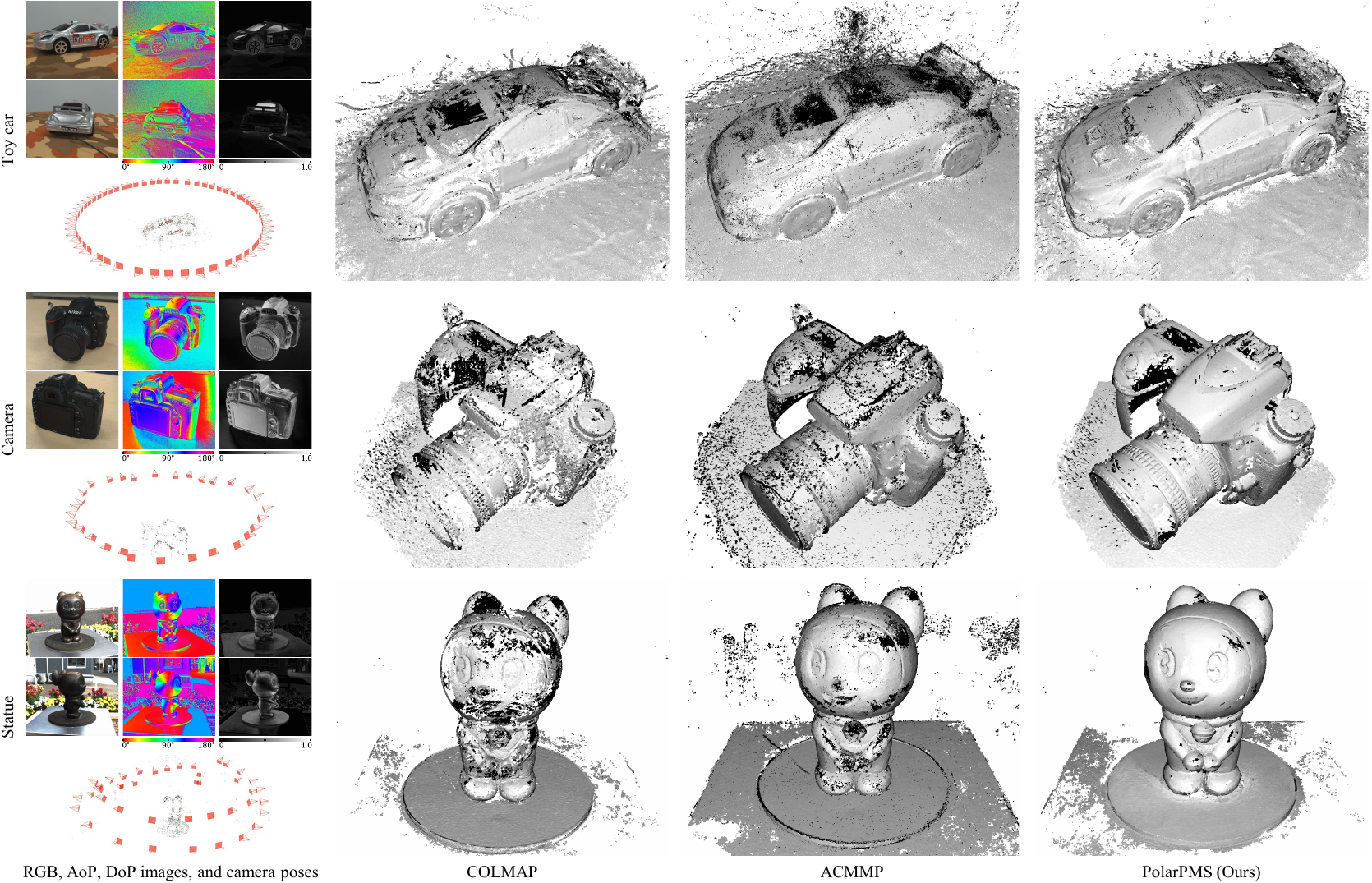}
   \caption{Examples of input images and the visual comparison using real data.}
   \label{fig:realScene_result}
\end{figure*}

We used the dataset in~\cite{zhao2022polarimetric} for a toy car~(56 views), a camera~(32 views), and a statue~(43 views). 
For real data, we empirically set $(\tau_{geo},\tau_{pol},\tau_{dep})$ to $(0.4,10.0,0.4)$ and $\rho_0$ to 1.0, seeing the actual strengths of noise and DoPs for real scenes. 
The results in Fig.~\ref{fig:realScene_result} show that COLMAP can generally reconstruct regions with relatively rich textures, but has limited performance in texture-less regions~(e.g., the front window of the toy car). Compared to COLMAP, ACMMP shows good performance in overall shape reconstruction, thanks to multi-scale geometry consistency to provide additional feature points and planar prior which can provide geometric constraints for the scene~(e.g., the base part of the statue). However, at the same time, the planar prior is principally difficult to apply for texture-less non-planar regions, since details of those regions are difficult to capture by fitting them to planes~(e.g., the side window of the toy car). In contrast, our PolarPMS can recover texture-less regions better~(e.g., the front window of the car and the face of the statue), and meantime reconstruct the details better~(the side window of the car, the lens part of the camera, and the bell of the statue). This demonstrates the effectiveness of our PolarPMS by utilizing polarimetric information for recovering texture-less regions and surface details.

\begin{figure}[t]
\includegraphics[width=1.0 \linewidth]{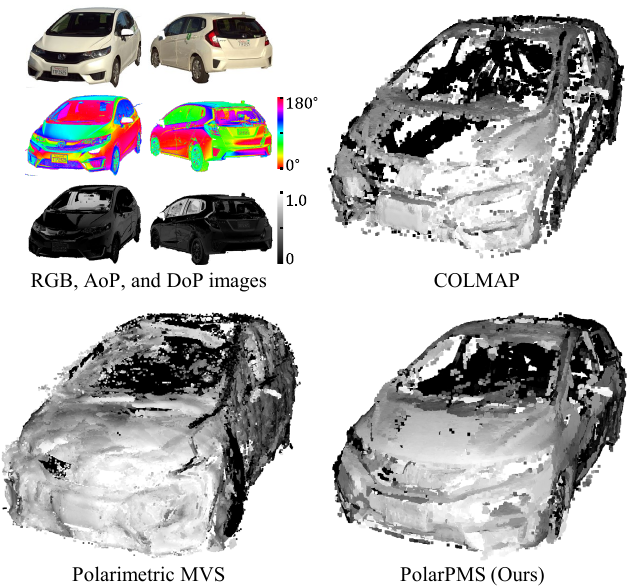}
   \caption{Examples of input images and the visual comparison for real car data provided by the authors of Polarimetric MVS.}
\label{fig:polarmvs}
\end{figure}

\subsection{Comparison with Polarimetric MVS}
We next compare PolarPMS with Polarimetric MVS~\cite{cui2017polarimetric}, which is a representative MVS method with polarization.
Since the source code of Polarimetric MVS is not available, we used the input and the result data provided by the authors of~\cite{cui2017polarimetric} for comparison. Figure~\ref{fig:polarmvs} shows the visual comparison for real car data. We can see that, compared to COLMAP which does not utilize polarization, both Polarimetric MVS and our PolarPMS reconstruct a much-completed 3D point cloud. While Polarimetric MVS tends to generate a very dense point cloud, it lacks accuracy and generates scattered 3D points for a single surface. This is because Polarimetric MVS uses AoP information for view-by-view depth propagation, but does not include multi-view consistency evaluation based on polarization. On the other hand, our PolarPMS generates more accurate 3D points by utilizing polarization information for multi-view consistency evaluation regarding the surface normal.

\section{Conclusion}
In this paper, we have proposed PolarPMS, which is a new PatchMatch MVS method that can improve the accuracy and completeness of reconstructed 3D models, compared with existing PatchMatch MVS methods. PolarPMS exploits polarization information to improve the estimation accuracy of the object's surface normal and depth, making use of the physical relationship between the observed AoP and the azimuth angle of the normal. Since this relationship is independent of the surface texture, PolarPMS realizes better completeness, especially for texture-less regions. Our PolarPMS has a limitation in reconstructing the regions where there exist strong inter-reflections, which have a more complicated status in polarization, and this encourages us to adopt a more accurate reflection model in our future work.

\section*{Acknowledgment}
This work was partly supported by JSPS KAKENHI Grant Number 21K17762. The authors would like to thank Dr. Zhaopeng Cui for sharing the data of Polarimetric MVS.

{\small
\bibliographystyle{ieee_fullname}
\bibliography{egbib}

\begin{thebibliography}{10}\itemsep=-1pt

\bibitem{Stanford}
\url{http://graphics.stanford.edu/data/3Dscanrep/}.

\bibitem{aanaes2016large}
Henrik Aan{\ae}s, Rasmus~Ramsb{\o}l Jensen, George Vogiatzis, Engin Tola, and
  Anders~Bjorholm Dahl.
\newblock Large-scale data for multiple-view stereopsis.
\newblock {\em Int. Journal of Computer Vision}, 120(2):153--168, 2016.

\bibitem{Agarwal_Ceres_Solver_2022}
Sameer Agarwal, Keir Mierle, and The Ceres~Solver Team.
\newblock {Ceres Solver}, 2023.

\bibitem{atkinson2007shape}
Gary~A Atkinson and Edwin~R Hancock.
\newblock Shape estimation using polarization and shading from two views.
\newblock {\em IEEE Trans. on Pattern Analysis and Machine Intelligence},
  29(11):2001--2017, 2007.

\bibitem{barnes2009patchmatch}
Connelly Barnes, Eli Shechtman, Adam Finkelstein, and Dan~B Goldman.
\newblock {PatchMatch}: A randomized correspondence algorithm for structural
  image editing.
\newblock {\em ACM Trans. on Graphics (TOG)}, 28(3):24, 2009.

\bibitem{bleyer2011patchmatch}
Michael Bleyer, Christoph Rhemann, and Carsten Rother.
\newblock Patchmatch stereo -- {Stereo} matching with slanted support windows.
\newblock In {\em Proc. of British Machine Vision Conference (BMVC)}, pages
  1--11, 2011.

\bibitem{cao2023multi}
Xu Cao, Hiroaki Santo, Fumio Okura, and Yasuyuki Matsushita.
\newblock Multi-view azimuth stereo via tangent space consistency.
\newblock In {\em Proc. of IEEE Conf. on Computer Vision and Pattern
  Recognition (CVPR)}, pages 825--834, 2023.

\bibitem{cui2017polarimetric}
Zhaopeng Cui, Jinwei Gu, Boxin Shi, Ping Tan, and Jan Kautz.
\newblock Polarimetric multi-view stereo.
\newblock In {\em Proc. of IEEE Conf. on Computer Vision and Pattern
  Recognition (CVPR)}, pages 1558--1567, 2017.

\bibitem{dave2022pandora}
Akshat Dave, Yongyi Zhao, and Ashok Veeraraghavan.
\newblock {PANDORA}: Polarization-aided neural decomposition of radiance.
\newblock In {\em Proc. of European Conf. on Computer Vision (ECCV)}, pages
  538--556, 2022.

\bibitem{duggal2019deeppruner}
Shivam Duggal, Shenlong Wang, Wei-Chiu Ma, Rui Hu, and Raquel Urtasun.
\newblock {DeepPruner}: Learning efficient stereo matching via differentiable
  patchmatch.
\newblock In {\em Proc. of IEEE Int. Conf. on Computer Vision (ICCV)}, pages
  4384--4393, 2019.

\bibitem{fukao2021polarimetric}
Yoshiki Fukao, Ryo Kawahara, Shohei Nobuhara, and Ko Nishino.
\newblock Polarimetric normal stereo.
\newblock In {\em Proc. of IEEE Conf. on Computer Vision and Pattern
  Recognition (CVPR)}, pages 682--690, 2021.

\bibitem{galliani2015massively}
Silvano Galliani, Katrin Lasinger, and Konrad Schindler.
\newblock Massively parallel multiview stereopsis by surface normal diffusion.
\newblock In {\em Proc. of IEEE Int. Conf. on Computer Vision (ICCV)}, pages
  873--881, 2015.

\bibitem{lee2021patchmatch}
Jae~Yong Lee, Joseph DeGol, Chuhang Zou, and Derek Hoiem.
\newblock {PatchMatch-RL}: Deep {MVS} with pixelwise depth, normal, and
  visibility.
\newblock In {\em Proc. of IEEE Int. Conf. on Computer Vision (ICCV)}, pages
  6158--6167, 2021.

\bibitem{ley2016syb3r}
Andreas Ley, Ronny H{\"a}nsch, and Olaf Hellwich.
\newblock {S}y{B}3{R}: A realistic synthetic benchmark for 3{D} reconstruction
  from images.
\newblock In {\em Proc. of European Conf. on Computer Vision (ECCV)}, pages
  236--251, 2016.

\bibitem{nimier2019mitsuba}
Merlin Nimier-David, Delio Vicini, Tizian Zeltner, and Wenzel Jakob.
\newblock Mitsuba 2: A retargetable forward and inverse renderer.
\newblock {\em ACM Trans. on Graphics (TOG)}, 38(6):1--17, 2019.

\bibitem{romanoni2019tapa}
Andrea Romanoni and Matteo Matteucci.
\newblock {TAPA-MVS}: Textureless-aware patchmatch multi-view stereo.
\newblock In {\em Proc. of IEEE Int. Conf. on Computer Vision (ICCV)}, pages
  10413--10422, 2019.

\bibitem{schoenberger2016sfm}
Johannes~Lutz Sch\"{o}nberger and Jan-Michael Frahm.
\newblock Structure-from-motion revisited.
\newblock In {\em Proc. of IEEE Conf. on Computer Vision and Pattern
  Recognition (CVPR)}, pages 4104--4113, 2016.

\bibitem{schoenberger2016mvs}
Johannes~Lutz Sch\"{o}nberger, Enliang Zheng, Marc Pollefeys, and Jan-Michael
  Frahm.
\newblock Pixelwise view selection for unstructured multi-view stereo.
\newblock In {\em Proc. of European Conf. on Computer Vision (ECCV)}, pages
  501--518, 2016.

\bibitem{smith2018height}
William Smith, Ravi Ramamoorthi, and Silvia Tozza.
\newblock Height-from-polarisation with unknown lighting or albedo.
\newblock {\em IEEE Trans. on Pattern Analysis and Machine Intelligence},
  41(12):2875--2888, 2019.

\bibitem{tian2023dps}
Chaoran Tian, Weihong Pan, Zimo Wang, Mao Mao, Guofeng Zhang, Hujun Bao, Ping
  Tan, and Zhaopeng Cui.
\newblock {DPS-Net}: Deep polarimetric stereo depth estimation.
\newblock In {\em Proc. of IEEE/CVF Int. Conf. on Computer Vision (ICCV)},
  pages 3569--3579, 2023.

\bibitem{wang2021patchmatchnet}
Fangjinhua Wang, Silvano Galliani, Christoph Vogel, Pablo Speciale, and Marc
  Pollefeys.
\newblock {PatchmatchNet}: Learned multi-view patchmatch stereo.
\newblock In {\em Proc. of IEEE Conf. on Computer Vision and Pattern
  Recognition (CVPR)}, pages 14194--14203, 2021.

\bibitem{wei2014multi}
Jian Wei, Benjamin Resch, and Hendrik~PA Lensch.
\newblock Multi-view depth map estimation with cross-view consistency.
\newblock In {\em Proc. of British Machine Vision Conference (BMVC)}, pages
  1--13, 2014.

\bibitem{xu2022multi}
Qingshan Xu, Weihang Kong, Wenbing Tao, and Marc Pollefeys.
\newblock Multi-scale geometric consistency guided and planar prior assisted
  multi-view stereo.
\newblock {\em IEEE Trans. on Pattern Analysis and Machine Intelligence},
  45(4):4945--4963, 2022.

\bibitem{xu2019multi}
Qingshan Xu and Wenbing Tao.
\newblock Multi-scale geometric consistency guided multi-view stereo.
\newblock In {\em Proc. of IEEE Conf. on Computer Vision and Pattern
  Recognition (CVPR)}, pages 5483--5492, 2019.

\bibitem{Xu2020ACMP}
Qingshan Xu and Wenbing Tao.
\newblock Planar prior assisted patchmatch multi-view stereo.
\newblock {\em AAAI Conf. on Artificial Intelligence (AAAI)},
  34(07):12516--12523, 2020.

\bibitem{yang2018polarimetric}
Luwei Yang, Feitong Tan, Ao Li, Zhaopeng Cui, Yasutaka Furukawa, and Ping Tan.
\newblock Polarimetric dense monocular {SLAM}.
\newblock In {\em Proc. of IEEE Conf. on Computer Vision and Pattern
  Recognition (CVPR)}, pages 3857--3866, 2018.

\bibitem{zhao2022polarimetric}
Jinyu Zhao, Yusuke Monno, and Masatoshi Okutomi.
\newblock Polarimetric multi-view inverse rendering.
\newblock {\em IEEE Trans. on Pattern Analysis and Machine Intelligence},
  45(7):8798--8812, 2023.

\bibitem{zheng2014patchmatch}
Enliang Zheng, Enrique Dunn, Vladimir Jojic, and Jan-Michael Frahm.
\newblock {PatchMatch} based joint view selection and depthmap estimation.
\newblock In {\em Proc. of IEEE Conf. on Computer Vision and Pattern
  Recognition (CVPR)}, pages 1510--1517, 2014.

\end{thebibliography}
}

\end{document}